\documentclass[10pt,twocolumn,letterpaper]{article}

\usepackage{cvpr}
\usepackage{times}
\usepackage{epsfig}
\usepackage{graphicx}
\usepackage{amsmath}
\usepackage{amssymb}
\usepackage{booktabs}
\usepackage[utf8x]{inputenc} 

\usepackage[breaklinks=true,bookmarks=false]{hyperref}

\cvprfinalcopy 

\def\cvprPaperID{4403} 

\ifcvprfinal\fi
\begin{document}

\title{Ensemble Generative Cleaning with Feedback Loops for Defending Adversarial Attacks}

\author{Jianhe Yuan and Zhihai He\\
University of Missouri, Columbia MO\\
{\tt\small \{yuanjia, hezhi\}@missouri.edu}
\and
}

\maketitle
\thispagestyle{empty}

\begin{abstract}
Effective defense of deep neural networks against  adversarial attacks remains a challenging problem, especially under powerful white-box attacks. In this paper, we develop a new method 
called \textit{ensemble generative cleaning with feedback loops} (EGC-FL) for effective defense of deep neural networks. The proposed EGC-FL method is based on two central ideas.
First, we introduce a transformed deadzone layer into the defense network, which consists of an orthonormal transform and a deadzone-based activation function,  to destroy the sophisticated noise pattern of adversarial attacks.
Second, by constructing a generative cleaning network with a feedback loop, we are able to generate an ensemble of diverse estimations of the original clean image. We then learn a network to fuse this set of diverse estimations together to restore the original image. Our extensive experimental results demonstrate that our approach improves the state-of-art by large margins in both white-box and black-box attacks. It significantly improves the classification accuracy for white-box PGD attacks upon the second best method by more than 29\% on the SVHN dataset and more than 39\% on the challenging CIFAR-10 dataset. 
\end{abstract}

\section{Introduction}

Researchers have recognized that deep neural networks are sensitive to adversarial attacks \cite{szegedy2013intriguing}. 
Very small changes of the input image can fool the state-of-art classifier with very high success probabilities. 
The attackers often generate noise patterns by exploiting the specific network architecture of the target deep neural network so that  small noise at the input layer can accumulate along the network inference layers, finally exceed the decision threshold at the output layer, and result in false decision. On the other hand, we know a well-trained deep neural networks are robust to random noise \cite{arjovsky2017wasserstein}, such as Gaussian noise. 
Therefore, the key issue in network defense is to destroy the sophisticated pattern or accumulative process of the attack noise while preserving the original image content or network classification performance.

During the past few years, a number of methods have been proposed to construct adversarial samples to attack the deep neural networks, including fast gradient sign (FGS) method \cite{goodfellow2014explaining}, Jacobian-based saliency map attack (J-BSMA) \cite{papernot2016limitations}, and projected gradient descent (PGD) attack \cite{kurakin2016adversarialscale,madry2018towards}. 
Different classifiers can be failed by the same adversarial attack method \cite{szegedy2013intriguing}. The fragility of deep neural networks and the availability of these powerful attacking methods present an urgent need for developing effective defense methods.
Meanwhile, deep neural network defense methods have also been developed, including adversarial training \cite{kurakin2016adversarialscale, szegedy2013intriguing}, defensive distillation \cite{papernot2016distillation, carlini2016defensive, papernot2016effectiveness}, Magnet \cite{meng2017magnet}, and featuring squeezing \cite{he2017adversarial, xu2017feature}.  
It has been recognized that these methods suffer from significant performance degradation under strong attacks, especially white-box attacks with large magnitude and iterations \cite{samangouei2018defense}. 

\begin{figure}
\begin{center}
 \includegraphics[width=\linewidth]{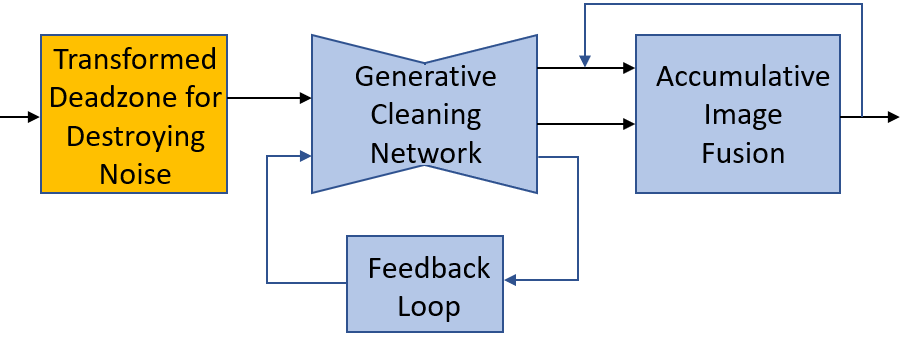}
\end{center}
\caption{Illustration of the proposed ensemble generative cleaning with feedback loops for defending adversarial attacks.}
\label{fig:idea}
\end{figure}

In this work, we explore a new approach, called \textit{ensemble generative cleaning with feedback loop} (EGC-FL), to defend deep neural network against powerful adversarial attacks.
Our approach is motivated by the following observation: (1) the adversarial attack  has sophisticated noise patterns which should be disturbed or destroyed during the defense process. (2) The attack noise, especially those powerful white-box attacks, such as the PGD and BPDA attacks \cite{ObfuscatedAthalye}, are often generated with an iterative process. To clean them, we also need an iterative process with multiple rounds of cleaning to achieve effective defense.

Motivated by these observations, our proposed EGC-FL approach first introduces a  transformed deadzone (TDZ) layer into the defense network, which consists of an orthonormal transform and a deadzone-based activation function, to destroy the sophisticated noise pattern of adversarial attacks.
Second, it introduces a new network structure with feedback loops, as illustrated in Figure \ref{fig:idea}, into the generative cleaning network. This feedback loop network allows us to remove the residual attack noise and recover the original image content in an iterative fashion. 
Specifically, over multiple feedback iterations, the EGC-FL network generates an  ensemble  of cleaned estimations of the original image.  Accordingly, we also learn an accumulative image fusion network which is able to fuse the new estimation with existing result in an iterative fashion.
According to our experiments, this feedback and iterative process converges very fast, often within 2 to 4 iterations.
Our extensive experimental results on benchmark datasets demonstrate that our EGC-FL approach improves the state-of-art by large margins in both white-box and black-box attacks. It significantly improves the classification accuracy for white-box attacks upon the second best method by more than 29\% on the SVHN dataset and more than 39\% on the challenging CIFAR-10 dataset with PGD attacks. 

The \textbf{major contributions} of this work can be summarized as follows. 
(1) We have introduced a transform deadzone layer into the defense network to effectively destroy the noise pattern of adversarial attacks.
(2) We have developed a new network structure with feedback loops to remove adversarial attack noise and recover original image content in an iterative manner.
(3) We have successfully learned an accumulative image fusion network which is able to fuse the incoming sequence of cleaned estimations and recover the original image in an iterative manner. 
(4) Our new method has significantly improved the performance of the state-of-the-art methods in the literature under a wide variety of attacks.

The rest of this paper is organized as follows. Section 2 reviews related work. The proposed EGC-FL method is presented in Section 3. Experimental results, performance comparisons with existing methods, and ablation studies are provided in Section 4. Section 5 concludes the paper.

\section{Related work}
\label{gen_inst}

In this section, we review related work on adversarial attack and network defense methods which are two tightly coupled research topics. The goal of attack algorithm design is to fail all existing network defense methods, while the goal of defense algorithms is to defend the deep neural networks against all existing adversarial attack methods.

\textbf{(A) Attack methods.} 
 Attack methods can be divided into two threat models: \textit{white-box} attacks and \textit{black-box} attacks. The white-box attacker has full access to the classifier network parameters, network architecture, and weights. The black-box attacker has no  knowledge of or access to the target network. 
 For white-box attack, a simple and fast approach called \textit{Fast Gradient Sign (FGS)} method has been developed by Goodfellow \textit{et al.} \cite{goodfellow2014explaining} using error back propagation to directly modify the original image. 
 Kurakin \textit{et al.} \cite{kurakin2016adversarialscale} apply FGS iteratively and propose BIM. Carlini \textit{et al.} \cite{carlini2016defensive} designed an optimization-based attack method, called \textit{Carlini-Wagner (C$\&$W) attack}, which is able to fool the target network with the smallest perturbation. 
 Xiao \textit{et al.} \cite{xiao2018generating} trained a generative adversarial network (GAN) \cite{goodfellow2014generative} to generate perturbations. 
 Kannan \textit{et al.} \cite{kannan2018adversarial} found that the \textit{Projected Gradient Descent (PGD)} is the strongest  among all attack methods. It can be viewed as a multi-step variant of FGS$^k$ \cite{madry2018towards}. Athalye \textit{et al.} \cite{ObfuscatedAthalye} introduced a method, called 
 Backward Pass Differentiable Approximation (BPDA), to attack networks where gradients are not available. It  iteratively computes the adversarial gradient on the defense results. It is able to successfully attack all existing state-of-the-arts defense methods. 
 For black-box attacks, the attacker has no knowledge about the target classifier. Papernot \textit{et al.} \cite{papernot2017practical} introduced the first approach for black-box attack using a substitute model. Dong \textit{et al.} \cite{Dong_2018} proposed a momentum-based iterative algorithms to improve the transferability of adversarial examples. Xie \textit{et al.} \cite{xie2018improving} boosted the transferability of adversarial examples by creating diverse input patterns.  

\begin{figure*}[t]
\begin{center}
 \includegraphics[width=\linewidth]{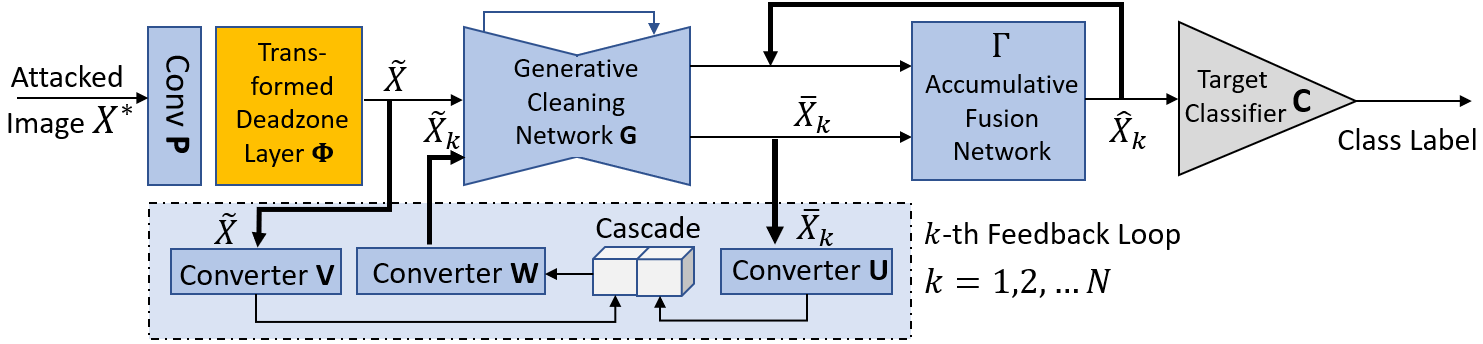}
\end{center}
\caption{Framework of the proposed ensemble generative cleaning network for defending adversarial attacks.}
\label{fig:framework}
\end{figure*}

\textbf{(B) Defense methods.}
Several approaches have recently been proposed for defending both white-box attacks and black-box attacks.  Adversarial training   trains the target model using adversarial examples  \cite{szegedy2013intriguing, goodfellow2014explaining}. Madry \textit{et al.} \cite{madry2018towards} suggested that training with adversarial examples generated by PGD improves the robustness. \cite{meng2017magnet} proposed a method, called MagNet, which detects the perturbations and then reshape them according to the difference between clean and adversarial examples. 
Recently, there are several defense methods based on GANs have been developed. Samangouei \textit{et al.} \cite{samangouei2018defense} projected the adversarial examples into a trained generative adversarial network (GAN) to approximate the input using generated clean image. 
Recently, some defense methods have been developed based on input transformations. Guo \textit{et al.} \cite{guo2018countering} proposed several input transformations to defend the adversarial examples, including image cropping and re-scaling, bit-depth reduction, and JPEG compression. Xie \textit{et al.} \cite{xie2018mitigating} proposed to defend against adversarial attacks by adding a randomization layer, which randomly re-scales the image and then randomly zero-pads the image. Jia \textit{et al.} \cite{Jia2018ComDefendAE} proposed an image compression framework to defend adversarial examples, called \textit{ComDefend}. Xie \textit{et al.} \cite{xie2018feature}  introduced a feature denoising method for defending PGD white-box attacks. 

\section{The Proposed Method}
\label{sec-overview}
In this section, we present our method of ensemble generative cleaning with feedback loops for defending adversarial attacks.

\subsection{Overview}

As illustrated in Figure \ref{fig:idea}, our proposed method of ensemble generative cleaning with feedback loops (EGC-FL) for defending adversarial attacks is based on \textbf{two main ideas}: (1) we introduce a transformed deadzone layer into the cleaning network to destroy the sophisticated noise patterns of adversarial attacks. (2) We introduce a generative cleaning network with a feedback loop  to generate a sequence of diverse estimations of the original image, which will be fused in an accumulative fashion to restore the original image. 

Figure \ref{fig:framework} shows a more detailed framework of the proposed EGC-FL method. The attacked image $X^*$ is first pre-processed by a convolutional layer $\mathbf{P}$ and then passed to the transformed deadzone layer $\mathbf{\Phi}$, which aims to destroy  the sophisticated noise patterns of the adversarial attacks. To remove the residual attack noise  in $\tilde{X}$ and recover the original image content $X$, the generative cleaning network $\mathbf{G}$ generates a series of estimations of the original image using a feedback loop. The feedback network consists of three converter networks, $\mathbf{U}$, $\mathbf{V}$, and $\mathbf{W}$, which are fully convolutional layers. 
These three converter networks are used to normalize the output features from different networks before they are concatenated or fused together.
At the $k$-th feedback loop, let $\bar{X}_k$ be the output of the generative cleaning network $\mathbf{G}$.  
We concatenate  the output $\bar{X}_k$ and the original $\tilde{X}$
after being normalized by converter networks $\mathbf{U}$ and $\mathbf{V}$, respectively. The concatenated feature map is  then normalized by  converter $\mathbf{W}$ before feeding back to the generative cleaning network $\mathbf{G}$ to produce the output 
$\bar{X}_{k+1}$. 
This feedback loop is summarized by the following formula:
\begin{equation}
    \bar{X}_{k+1} = \mathbf{G}\{\mathbf{W}[\mathbf{V}(\tilde{X}) \uplus \mathbf{U}(\bar{X}_{k}) ]\},
\end{equation}
where $\uplus$ represents the cascade operation. 
This ensemble generative cleaning network with feedback will generate a series of cleaned versions $\{\bar{X}_k|k=1,2,\cdots \}$, representing a diverse set of estimations of the original image $X$. To recover the original image $X$, we introduce an accumulative image fusion network $\mathbf{\Gamma}$, which operates as follows
\begin{equation}
    \hat{X}_{k+1} = \mathbf{\Gamma}(\hat{X}_{k}, \bar{X}_k).
\end{equation}
Specifically, the input to the fusion network $\mathbf{\Gamma}$ are two images:
$\bar{X}_k$ which is the current output from the generative cleaning network $\mathbf{G}$, and $\hat{X}_{k}$ which is the current fused image produced by  $\mathbf{\Gamma}$. 
The generative cleaning network $\mathbf{G}$ is separated from the accumulative fusion network $\mathbf{\Gamma}$ so that the generative network can generate multiple estimations of the original image. The fusion network can then fuse them together.
In other words, the output of $\mathbf{\Gamma}$ is fed back to itself as the input for the next round of fusion. All networks, including the convolution pre-processing, the generative cleaning network, converter networks, and the accumulative fusion network are learned from our training data, which will be explained in more detail in the following sections.

\begin{figure}
\begin{center}
 \includegraphics[width=0.55\linewidth]{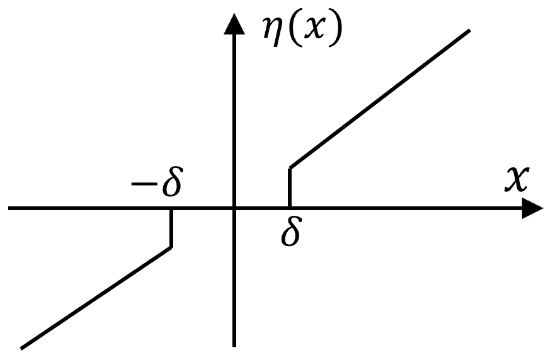}
\end{center}
\caption{Activation function for the TDZ layer.}
\label{fig:tdz}
\end{figure}

\subsection{Transformed Deadzone Layer}
The goal of the transformed deadzone layer in our defense network is to destroy the noise pattern and perform the first round of removal of the adversarial attack noise. 
Let $X$ be the original image and $x(i,j)$ be its pixel at location $(i, j)$. The attacked image is given by $X^* = X + \mathbf{\alpha}_\epsilon $ 
where $\mathbf{\alpha}_\epsilon$ is adversarial attack with magnitude $\epsilon$ and  $\alpha_\epsilon(i,j)$ is the attack noise at pixel location $(i, j)$, which is a random variable with maximum magnitude of $\epsilon$. We have $x^*(i,j)=x(i,j)+\alpha_\epsilon(i,j)$. In the spatial domain, it is very challenging  to separate the attack noise from the original image content since the attacked image $X^*$ and the original image $X$ are visually very similar to each other perceptually.

\begin{figure}
\begin{center}
  \includegraphics[width=\linewidth]{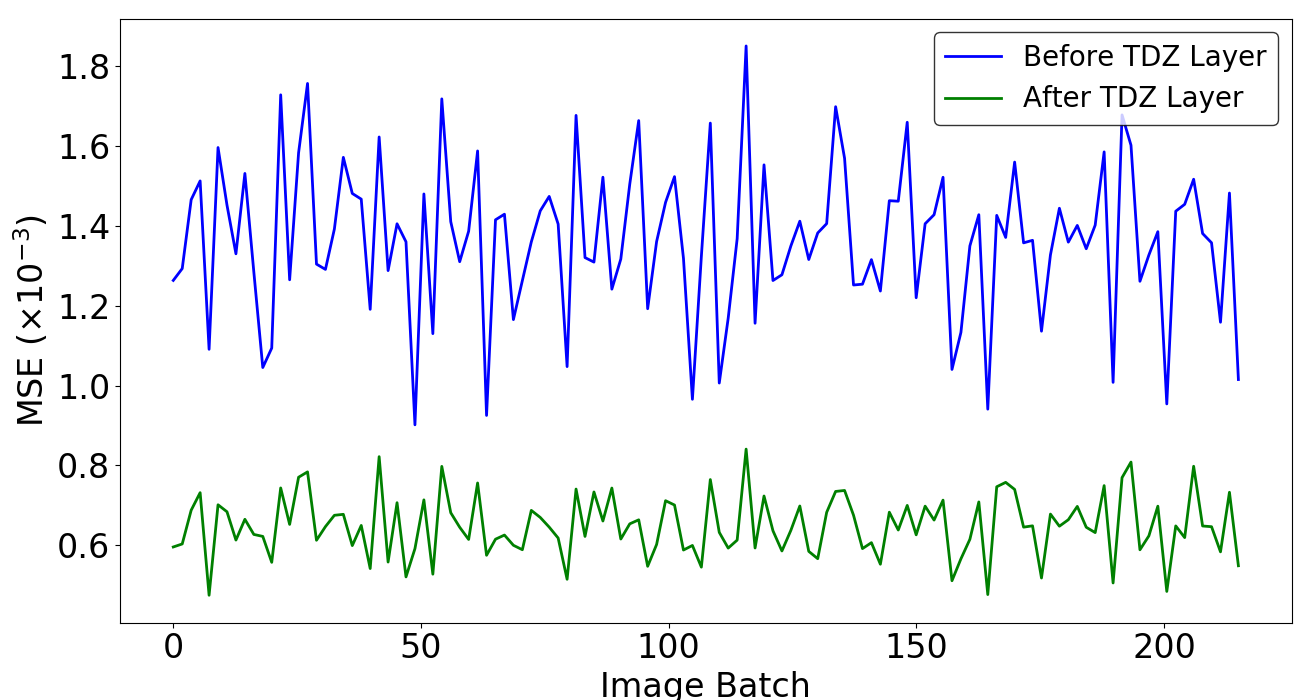}
\end{center}
\caption{The energy of attack noise before and after the transformed TDZ for 860 test images from 215 batches.}
\label{fig:tdz-results}
\end{figure}

To address this issue, we propose to first transform the image using a de-correlation or energy compaction orthonormal transform matrix $\mathbf{T}$. One choice of this transform is the blockwise discrete cosine transform (DCT) \cite{wallace1992jpeg}. After this transform, the energy of the original image will be aggregated onto a small fraction of transform coefficients with the remaining coefficients being very close to zeros. We then pass this transformed image through a deadzone activation function $\eta(x)$ shown in Figure \ref{fig:tdz}. Here, $\eta(x) = 0$ if $x\in [-\delta, \delta]$. Otherwise $\eta(x) = x$.
Since the transform is linear, the transformed image after the deadzone activation is given by 
\begin{eqnarray}
\eta(T\cdot X^*) &=& \eta(T\cdot X + T\cdot \mathbf{\alpha}_\epsilon), \\ 
\eta(x_t^*(i, j)) &=& \eta(x_t(i, j)) + \eta(\alpha_t(i, j)) \\
&\approx& \eta(x_t(i, j)).
\end{eqnarray}
Statistically, the attack noise is white noise. After transform,  $\alpha_t(i, j)$ remains white noise.
Notice that a vast majority of transform coefficients $x_t(i, j)$ in the transformed image 
$T\cdot X$ will be very small. In this case, the deadzone activation function $\eta(x)$ will largely remove the transformed attack noise $\alpha_t(i, j)$. Meanwhile, since the major image content or energy has be aggregated onto a smaller number large-valued coefficients, which remain unchanged by the deadzone function. In this way, the energy-compaction transform is able to help protecting the original image content from being damaged by the deadzone activation function during removal of attack noise. 
Certainly, it will still cause some damage to the original image content since the small transform coefficients $x_t(i, j)$ are forced to zeros. 
Figure \ref{fig:tdz-results} shows the energy of the attack noise  before and after the TDZ, namely,  $||\alpha_\epsilon||_2 = ||X - X^*||_2$ 
and  
$||\Gamma(X) - \Gamma(X^*)||_2$,  for 860 test images organized in 215 batches. Here 
$\Gamma(\cdot)$ represents the transformed deadzone operation. We can see that the energy of attack noise has been significantly reduced. Certainly, some parts of the original image content, especially those high-frequency details, are also removed, which need to be recovered by the subsequent generative cleaning network.

\begin{table*}
  \caption{Performance of our method (classification accuracy after defense) against white-box attacks on CIFAR-10 dataset ($\epsilon$ = $8/256$). Some methods did not provide results on specific attack methods, which were left blank (marked with \scriptsize{(NA)} \normalsize{).}}
  \label{cifar10-table-white-c}
  \centering
  \begin{tabular}{c|c|ccccc}
    \toprule
    Defense Methods & Clean & FGS & PGD &BIM &C\&W \\
    \midrule
    \textit{No Defense} & \textit{94.38\%}  & \textit{31.89\%} & \textit{0.00\%} &\textit{0.00\%} &  \textit{0.99\%} \\

    Label Smoothing \cite{warde201611} & 92.00\% & 54.00\% & \scriptsize{(NA)}  & 8.00\% & 2.00\%\\
    Feature Squeezing \cite{xu2017feature}& 84.00\% & 20.00\% & \scriptsize{(NA)}  & 0.00\% & 78.00\%\\
    PixelDefend \cite{song2018pixeldefend} & 85.00\% & 70.00\% & \scriptsize{(NA)}  & 70.00\% & 80.00\%\\
    Adv. Network \cite{wang2018aDirect} & 91.08\% & 72.81\%  &  44.28\%  & \scriptsize{(NA)}& \scriptsize{(NA)}    \\
    Parametric Noise Injection (PNI)  \cite{he2019parametric} & 85.17\%& 56.51\% & 49.07\% & \scriptsize{(NA)}& \scriptsize{(NA)}\\
     Sparse Transformation Layer (STL) \cite{Sun_Tsai_Liu_Yu_Su_2019} & 90.11\% & 87.15\%& \scriptsize{(NA)}& 88.03\%& 89.04\% \\
    \midrule
    {Our Method} & {91.65\%} & \textbf{88.51\%} & \textbf{88.61\%} & \textbf{88.75\%}& \textbf{90.03\%} \\
    Gain &  & +1.36\% & +39.54\% & +0.72\% & +0.99\% \\
    \bottomrule
  \end{tabular}
\end{table*}

\subsection{Learning the Ensemble Generative Cleaning Network}
In our defense method design, the generative cleaning network $\mathbf{G}$, the feedback loop $\mathbf{U,V,W}$ and accumulative fusion network $\mathbf{\Gamma}$ are jointly trained. 
The goal of our method is three-fold: (1) first, the generative cleaning network $\mathbf{G}$ needs to make sure that the original image content is largely recovered. 
(2) Second, the feedback loop  needs to successfully remove the residual attack noise.
(3) Third, the accumulative fusion network $\mathbf{\Gamma}$ needs to iteratively recover the original image content.
To achieve the above three goals, we formulate the following generative loss function for training the networks
 \begin{equation}
     \mathbf{L} =\lambda_1 \mathbf{L}_{P} + \lambda_2 \mathbf{L}_{A} + \lambda_3 \mathbf{L}_{C},
 \end{equation}
where $\mathbf{L}_P$ is perceptual loss, $\mathbf{L}_A$ is the adversarial loss and $\mathbf{L}_C$ is the cross-entropy loss. $\lambda_i$ is a weighting parameter. In our experiments, we set it to be 1/3. 
To define the perceptual loss, the $L_2$-norm between the recovered image $\hat{X}_{k}$ and the original image $X$ is used \cite{johnson2016perceptual}. In this work, we observe that the small adversarial perturbation often leads to very substantial noise in the feature map of the network \cite{xie2018feature}. Motivated by this, we use a pre-trained VGG-19 network, denoted by $\mathbf{F}_\beta$ to generate visual features for the recovered image $\hat{X}_{k}$ and the original image $X$, and use their feature difference as the perceptual loss $\mathbf{L}_P$. Specifically, 
\begin{equation}
     \mathbf{L}_P = ||\mathbf{F}_\beta(X)- \mathbf{F}_\beta(\hat{X}_{k}) ||_2^2.
\end{equation}
The adversarial loss $\mathbf{L}_A$ aims to train  generative cleaning network $\mathbf{G}$ and the feedback loop $\mathbf{U,V,W}$ so that the recovered images will be correctly classified by the target network. It is formulated as 
\begin{equation}
    \mathbf{L}_A =  ||\mathbf{G}\{\mathbf{W}[\mathbf{V}(\tilde{X}) \uplus \mathbf{U}(\bar{X}_{k})]\} - X||_2^2.
\end{equation}
We train our accumulative fusion network $\mathbf{\Gamma}$, along with the generative cleaning network $\mathbf{G}$, to optimize the following loss function:
\begin{equation}
    \mathbf{L}_C = \mathbb{E}_{{X}\in \Omega}\Phi [\mathbf{\Gamma}(\hat{X}_{k}, \bar{X}_k), \mathbf{I}_{clean}].
\end{equation}
Here, $\Phi[\cdot, \cdot]$ represents the cross-entropy between the output generated by the generative network and the target label $ \mathbf{I}_{clean}$ for clean images. 
With the above loss functions, our ensemble generative cleaning network learns to iteratively recover adversarial images.

The accumulative fusion network $\mathbf{\Gamma}$ acts as a multi-image restoration network for original image reconstruction. 
Cascaded with the generative cleaning network $\mathbf{G}$, it will guide the training of $\mathbf{G}$ and feedback loop network using back propagation of gradients from its own network, aiming to minimize the above loss function. 
In our design, during the adversarial learning process, the target classifier $\mathbf{C}$ is called to determine if the recovered image $\hat{X_k}$ is clean or not, as illustrated in Figure \ref{fig:framework}.  The output of $\mathbf{\Gamma}$ is fed back to itself as the input to enhance the next round of fusion.

\section{Experimental Results}
In this section, we implement and evaluate our EGC-FL  defense method and compare its performance with state-of-the-art defense methods under a wide variety of attacks, with both white-box and black-box attack modes.

\subsection{Experimental Settings}
Our experiments are implemented on the Pytorch
platform \cite{paszke2017automatic}. Our proposed method is implemented on the AdverTorch \cite{ding2018advertorch} in both white and black-box attack modes, including the BPDA attack \cite{ObfuscatedAthalye}. 
We choose the CIFAR-10 and SVHN (Street View House Number) datasets for performance evaluations and comparisons since most recent papers reported results on these two datasets. 
The CIFAR-10 dataset consists of 60,000 images in 10 classes of size $32\times32$. The Street View House Numbers (SVHN) dataset \cite{NetzerReading} has about 200K images of street numbers. For each of these two datasets, a classifier is independently
trained on its training set, and the test set is used for evaluations.

\begin{table}
  \caption{ BPDA attack results on CIFAR-10 dataset. Results with  $∗$ are achieved with additional adversarial training. }
  \label{cifar10-table-white-BPDA}
  \centering
  \begin{tabular}{c|c}
    \toprule
     Defense Methods  &  Accuracy \\
    \midrule
    Thermometer Encodings (TE) \cite{buckman2018thermometer}  & ${0.00\%}^*$ \\
    Stochastic Activation Pruning (SAP) \cite{2018stochastic}  & 0.00\% \\
    Local Intrinsic Dimensionality (LID) \cite{ma2018characterizing} & 5.00\% \\
    PixelDefend \cite{song2018pixeldefend} & ${9.00\%}^*$ \\
    Cascade Adv. Training ($L_\infty$=0.015) \cite{na2018cascade} & ${15.00\%}$ \\
    PGD Adv. Training \cite{madry2018towards} & ${47.00\%}^*$ \\
     Sparse Transformation Layer (STL) \cite{Sun_Tsai_Liu_Yu_Su_2019} & ${42.00\%}^*$ \\
    \midrule
    {Our Method }  & $\textbf{85.77\%}^*$ \\
    Gain & +38.77\%\\
    \bottomrule
  \end{tabular}
\end{table}

\begin{table}
  \caption{Performance of our method against black-box attacks on CIFAR-10 ($\epsilon$ = $8/256$).}
  \label{cifar10-table-black}
  \centering
  \begin{tabular}{c|c|cc}
    \toprule
     Defense Methods & No Attack & FGS & PGD   \\
    \midrule
    \textit{No Defense} & \textit{94.38\%}   & \textit{63.21\%} & \textit{38.71\%}    \\
    Adv. PGD \cite{tramr2018ensemble} & 83.50\% &57.73\% & 55.72\%   \\
    Adv. Network \cite{wang2018aDirect} & 91.32\% & 77.23\% & 74.04\% \\
    \midrule
    {Our Method} & {91.65\%} & \textbf{79.09\%} & \textbf{82.78\%} \\
    Gain &  & +1.86\% & +8.74\%\\
    
    \bottomrule
  \end{tabular}
\end{table}

\subsection{Results on the CIFAR-10 Dataset}

We compare the performance of our defense method with state-of-the-art methods developed in the literature under five different white-box attacks: (1) FGS attack \cite{goodfellow2014explaining}, (2) PGD attack \cite{madry2018towards}, (3) BIM attack \cite{kurakin2016adversarialscale}, (4) C\&W attack \cite{Carlini_2017}, and (5) BPDA attack \cite{ObfuscatedAthalye}. 
Following \cite{kannan2018adversarial} and \cite{wang2018aDirect}, the white-box attackers generate adversarial perturbations within a range of $\mathbf{\epsilon}=8/255$. In addition, we set the step size of attackers to be $\mathbf{\epsilon}=1/255$ with $10$ attack iterations as the baseline setting.

We generate the perturbed images for training using PGD attacks and tested for all attack methods. During training, we set the iteration number $K=3$. The perturbed images are used as the input, passing through our EGC-FL network for 3 iterations. But, during test, $K$ is flexible. In our white-box attack experiments, we unfold the feedback loops so that the  attacker has full access to the end-to-end defense network, including the number of iterations.

\textbf{(1) Defending against white-box attacks.} 
Table \ref{cifar10-table-white-c} shows image classification accuracy with 6 defense methods: (1) Label Smoothing \cite{warde201611}, (2) Feature Squeezing \cite{xu2017feature}, (3) PixelDefend \cite{song2018pixeldefend}, (4) Adversarial Network \cite{wang2018aDirect}, (5) the PNI (Parametric Noise Injection) method \cite{he2019parametric}, and (6) the STL (Sparse Transformation Layer) method \cite{Sun_Tsai_Liu_Yu_Su_2019}. The second column shows the classification accuracy when the input images are all clean. 
We can see that some methods, such as the PixelDefend \cite{song2018pixeldefend}, Feature Squeezing \cite{xu2017feature}, and PNI \cite{he2019parametric}, degrade the classification accuracy of clean images. This implies that their defense methods have caused significant damages to the original images, or they cannot accurately tell if the input image is clean or being attacked.
Since our method has a strong reconstruction capacity, the ensemble of reconstructed images still preserve the useful information. 
The rest four columns list the final image classification accuracy with different defense methods. 
For all of these four attacks, our methods significantly outperforms existing methods. For example, for the powerful PGD attack, our method outperforms the Adv. Network and the PNI  method by more than 39\%.

\textbf{(2) Defending against the BPDA attack.} The Backward Pass Differentiable Approximation (BPDA) \cite{ObfuscatedAthalye} attack is very challenging to defend since it can iteratively strengthen the adversarial examples using gradient approximation according to the defense mechanism.
BPDA also targets defenses in which the gradient does not optimize the loss. This is the case for our method since the transformed deadzone layer is non-differentiable.
Table \ref{cifar10-table-white-BPDA} summarizes the defense results of our algorithm in comparison with other seven methods: (1) Thermometer Encodings (TE) \cite{buckman2018thermometer}, (2) Stochastic Activation Pruning (SAP) \cite{2018stochastic}, (3) Local Intrinsic Dimensionality (LID) \cite{ma2018characterizing}, (4) PixelDefend \cite{song2018pixeldefend}, (5) Cascade Adversarial Training \cite{na2018cascade}, (6) PGD Adversarial Training \cite{madry2018towards}, and (7) Sparse Transformation Layer (STL) \cite{Sun_Tsai_Liu_Yu_Su_2019}. We choose these methods for comparison since the original BPDA paper \cite{ObfuscatedAthalye} has reported results of these methods.
We can see that our EGC-FL network is much more robust than other defense methods on the CIFAR-10 dataset,
outperforming the second best by more than 38\%.


\begin{table}
  \caption{Performance of our method against white-box attacks on SVHN ($\epsilon$ = $12/256$).}
  \label{svhn-table-white}
  \centering
  \begin{tabular}{c|c|cc}
    \toprule
     Defense Methods & No Attack & FGS & PGD  \\
    \midrule
    \textit{No Defense} & \textit{96.21\%}  & \textit{50.36\%}  & \textit{0.15\%}   \\
    M-PGD \cite{madry2018towards} & 96.21\% & \scriptsize{(NA)} & 44.40\%       \\
    ALP \cite{kannan2018adversarial} & 96.20\% & \scriptsize{(NA)} & 46.90\%      \\
    Adv. PGD \cite{tramr2018ensemble} & 87.45\% & 55.94\%  & 42.96\%     \\
    Adv. Network \cite{wang2018aDirect} & 96.21\% & 91.51\%  & 37.97\%    \\
    \midrule
    Our Method  & 94.00\% & \textbf{94.10\%} & \textbf{76.67\%}  \\
    Gain &  & +2.59\% & +29.77\%\\
    \bottomrule
  \end{tabular}
\end{table}


\textbf{(3) Defending against black-box attacks.} 
We generate the black-box adversarial examples  using FGS and PGD attacks with a substitute model \cite{papernot2017practical}. 
The substitute model is trained in the same way as the target classifier with a ResNet-34 network \cite{He_2016_CVPR} structure.
Table \ref{cifar10-table-black} shows the performance of our defense mechanism under back-box attacks on the CIFAR-10 dataset. 
The adversarial examples are constructed with $\epsilon$ = $8/256$ under the substitute model.
We observe that the target classifier is much less sensitive to adversarial examples generated by FGS and PGD black-box attacks than the white-box ones. But the powerful PGD attack is still able to decrease the overall classification accuracy to a very low level, $38.71\%$. 
We compare our method with the Adversarial PGD \cite{madry2018towards} and Adversarial Network \cite{wang2018aDirect} methods. We include these two because they are the only ones that provide performance results on CIFAR-10 with black-box attacks.
From the Table \ref{cifar10-table-black}, we can see our method improves the accuracy by 8.74\% over the state-of-the-art Adversarial Network method for the PGD attack.

\subsection{Results on the SVHN Dataset.} 

We evaluate our EGC-FL method on the SVHN dataset with comparison with four state-of-the-art defense methods: (1) M-PGD (Mixed-minibatch PGD ) \cite{madry2018towards}, (2) ALP (Adversarial Logit Pairing) \cite{kannan2018adversarial}, (3) Adversarial PGD \cite{tramr2018ensemble}, and (4) Adversarial Network \cite{wang2018aDirect}. 
For the SVHN dataset, as in the existing methods \cite{kannan2018adversarial, wang2018aDirect}, we used the Resnet-18 \cite{He_2016_CVPR}  for the  target classifier. The average classification accuracy is $96.21\%$. We use the same parameters as in  \cite{kannan2018adversarial} for the PGD attack with a total magnitude of $\epsilon=0.05\ (12/255)$. Within each single step, the  perturbation magnitude is set to be  $\epsilon=0.01\ (3/255)$ and  $10$ iterative steps are used.

\textbf{(1) Defending against white-box attacks.} 
Table \ref{svhn-table-white} summarizes the experimental results and performance comparisons with those four existing defense methods. We can see that on this dataset the PGD attack is able to decrease the overall classification accuracy to an extremely low level, 0.15\%. 
Our algorithm outperforms existing methods by a very large margin. For example,  for the PGD attack, our algorithm outperforms the second best ALP \cite{kannan2018adversarial} algorithm by more than $29\%$. 
With the FGS attacks, the iterative cleaning process will produce image versions with more diversity than the clean image without attack noise. This helps reconstruct the original image.

\begin{table}
  \caption{Performance of our method against black-box attacks on SVHN ($\epsilon$ = $12/256$).}
  \label{svhn-table-black}
  \centering
  \begin{tabular}{c|c|cc}
    \toprule
     Defense Methods & No Attack & FGS & PGD  \\
    \midrule
    \textit{No Defense} & \textit{96.21\%}  & \textit{69.91\%} & \textit{67.66\%}    \\
    M-PGD \cite{madry2018towards} & 96.21\% & \scriptsize{(NA)} & 55.40\%       \\
    ALP \cite{kannan2018adversarial} & 96.20\% & \scriptsize{(NA)} & 56.20\%      \\
    Adv. PGD \cite{tramr2018ensemble} & 87.45\% & 87.41\%  & 83.23\%     \\
    Adv. Network \cite{wang2018aDirect} & 96.21\% & 91.48\%  & 81.68\%    \\
    \midrule
    Our Method & 94.00\% & \textbf{94.03\%} & \textbf{88.60\%}\\
    Gain &  & +2.55\% & +5.37\%\\
    \bottomrule
  \end{tabular}
\end{table}

\begin{figure}
\begin{center}
  \includegraphics[width=\linewidth]{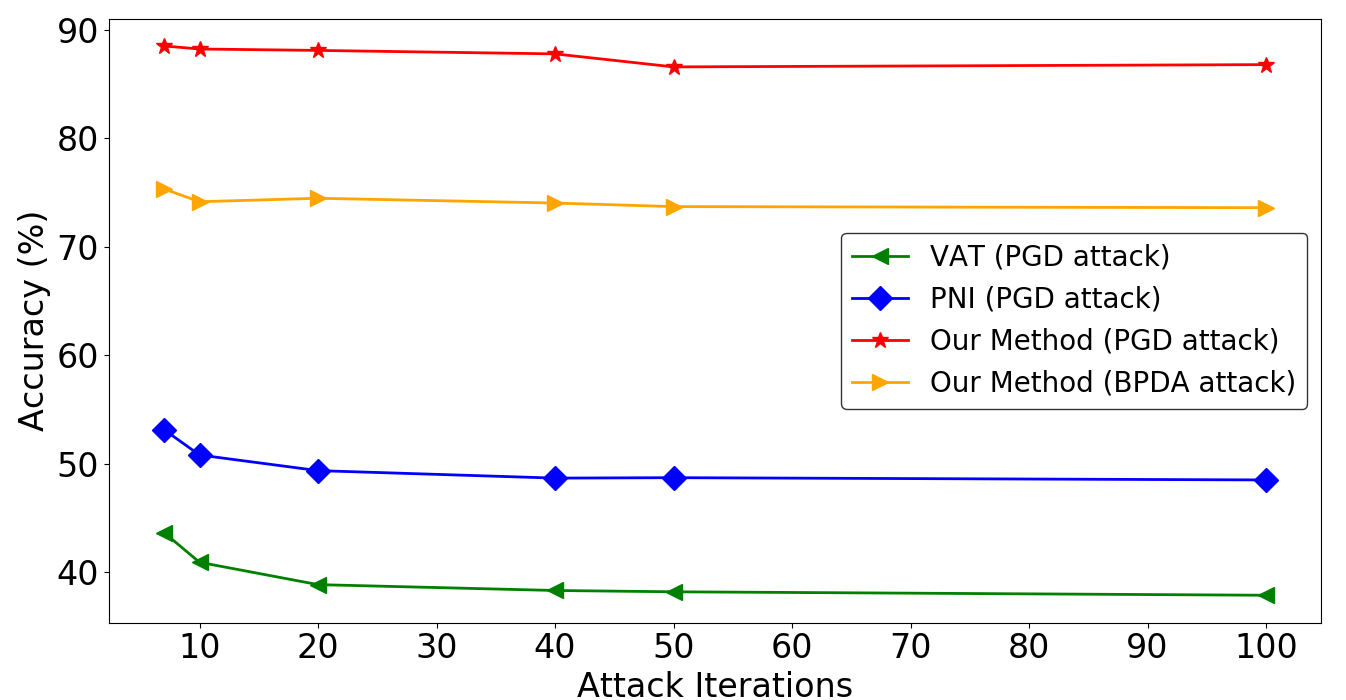}
  \includegraphics[width=\linewidth]{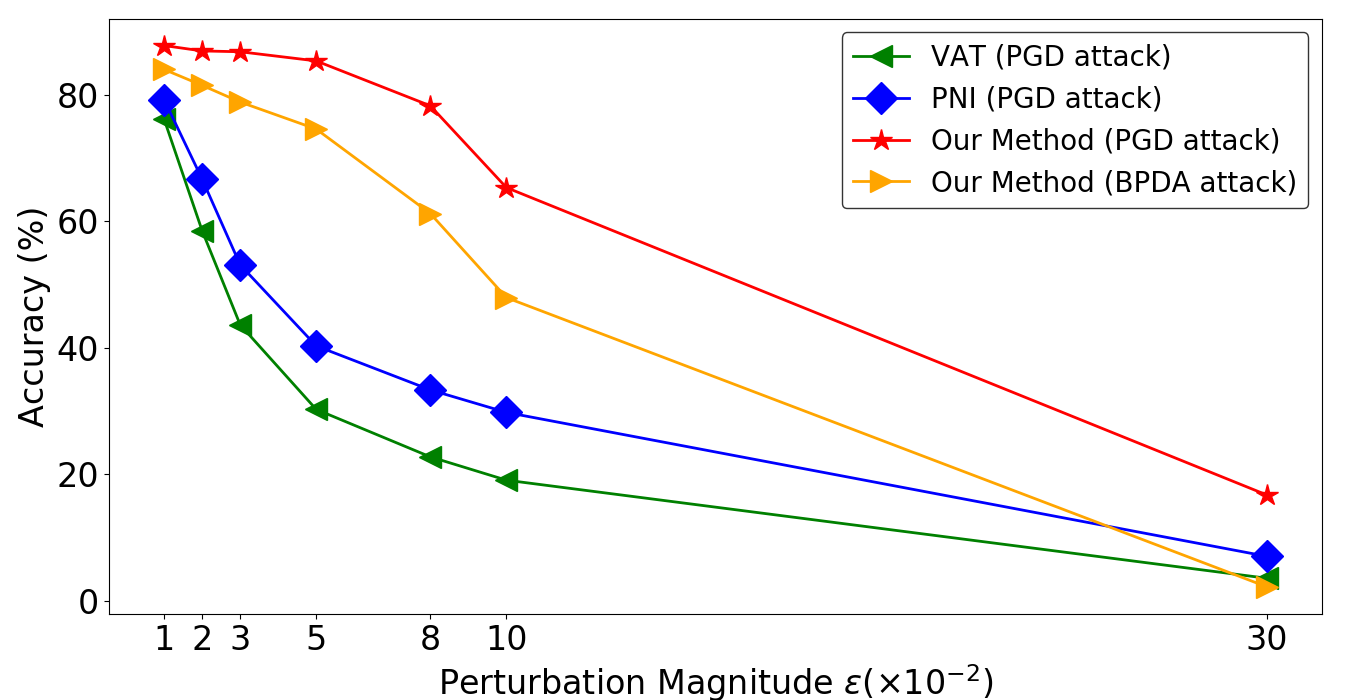}
\end{center}
\caption{ The perturbed-data accuracy of ResNet-18 under adversarial attack (Top) versus number of attack iteration, and (Bottom) versus perturbation magnitude (under $L_\infty$) on CIFAR-10 dataset.}
\label{fig:iter_attack}
\end{figure}

\textbf{(2) Defending against black-box attacks.} We also perform experiments of defending black-box attacks on the SVHN dataset.
Table \ref{svhn-table-black} summarizes our experimental results with the powerful PGD attack and provides the comparison with those four methods. We can see that our approach outperforms other methods by $2.25\%$ for the FGS attacks and $5.37\%$ for the PGD attacks.
From the above results, we can see that our proposed method is particularly effective for defense against the strong attacks, for example, the PGD attacks with large iteration steps and noise magnitude. 

\subsection{Ablation Studies and Algorithm Analysis}

In this section, we provide in-depth ablation study results of our algorithm to further understand its capability. 

\textbf{(1) Defense against large-iteration and large-epsilon attacks.}
Figure \ref{fig:iter_attack} (Top) shows the performance results under large-iteration PGD and BPDA attacks. 
We can see that the large-iteration PGD attack significantly degrades the accuracy of the Vanilla Adversary Training method (VAT) \cite{madry2018towards} and the PNI (Parametric Noise Injection) method \cite{he2019parametric}, as well as our method. 
But, our method significantly outperforms the other two.
In both cases, the perturbed-data accuracy starts saturating without further drop  when $N_{step}\ge 50$.
In Figure \ref{fig:iter_attack} (Top), we also include the performance results of our method under large-iteration BPDA attacks. 
We set the adversarial perturbations within a range of $\mathbf{\epsilon}=12/255$ with $10$ attack iterations as the baseline setting.
This result is not reported by other methods so we could not include them for comparison. We can see that the BPDA attack is much more powerful. But, our algorithm can still survive large-iteration BPDA attacks and largely maintain the defense performance.

Figure \ref{fig:iter_attack} (bottom) shows comparison results against attacks with large perturbation magnitude.  
We can see that our method significantly outperforms the VAT and PNI defense methods even when the magnitude of adversarial noise is increased  to $\epsilon=0.3$ under the PGD attack. 
We also include the performance of our method under large-$\epsilon$ BPDA attacks. 
We can see  that our method is robust under very powerful attacks of large magnitudes.

\begin{figure}
\begin{center}
  \includegraphics[width=\linewidth]{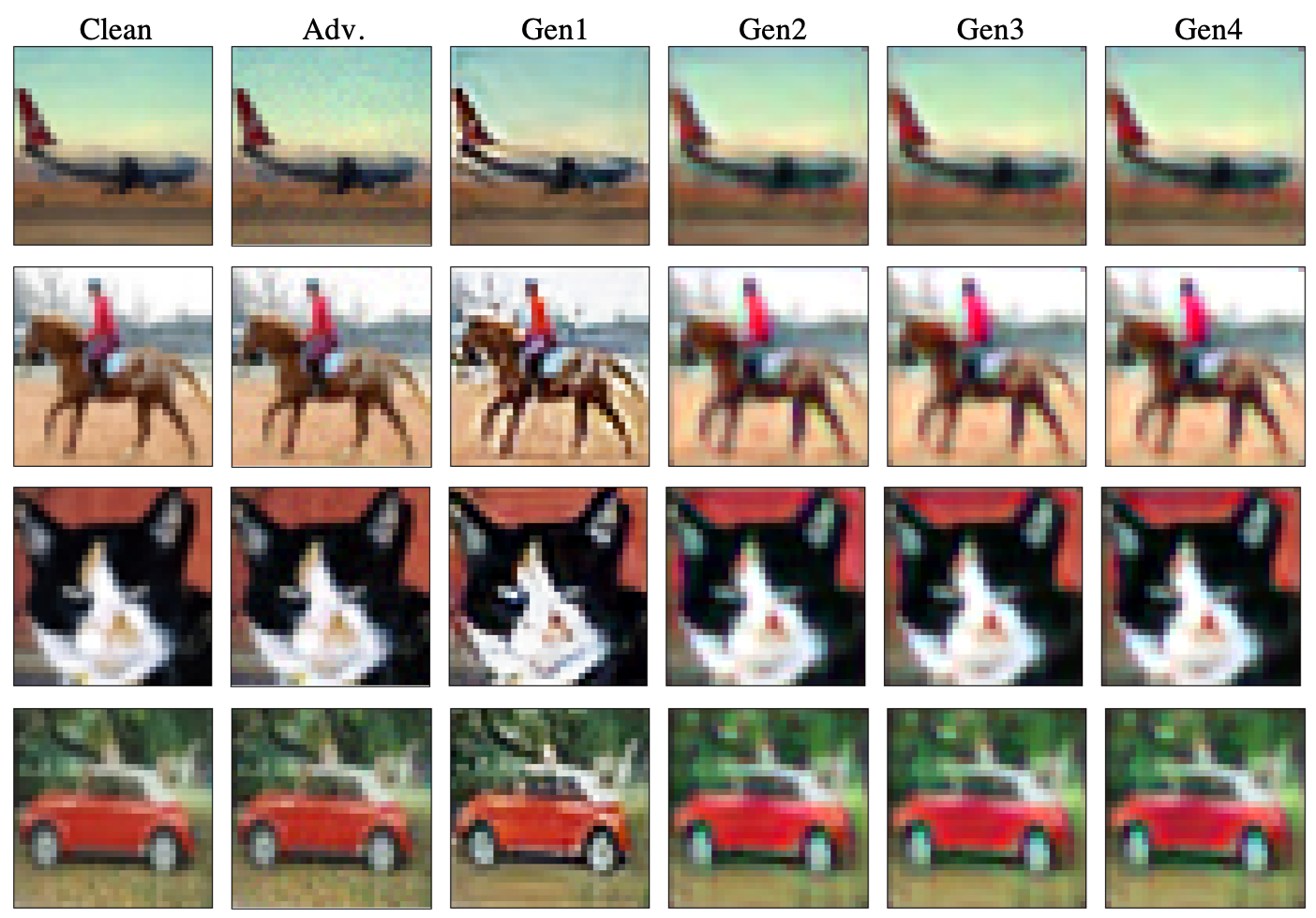}
\end{center}
\caption{Adversarial images and their fused image produced by our method.}
\label{fig:adv_loop}
\end{figure}

\textbf{(2) Analyze the impact of feedback loops.}
We notice that the feedback loop network plays an important role in the defense. 
In our method, the key parameter controlling the image quality is the number of feedback loop $k$. We gradually increase $k$ and explore classification accuracy of the fused image. 
Table \ref{feedback_loop} shows the performance (classification accuracy after defense) of our method on the CIFAR-10 dataset with various attacks.
We denote Gen$_k$ as the number of feedback loops. We can see that the feedback loops within the range of 3 or 4 yields the best performance. One feedback loop does not provide efficient defense since the EGC-FL network is not able to fully destroy the attack noise pattern and restore useful information. Once the key features in the original image have been reconstructed, the classification accuracy will be stable and maintain the highest performance, although the image quality may get even better with accumulative fusion. 
In Figure \ref{fig:adv_loop}, we show sample images from the CIFAR-10 when our method is applied. The first column is the clean image without attacks. The second column is attacked image. The third to last columns are reconstructed images of 4 generations by our EGC-FL method.  We can see that our algorithm is able to remove the attack noise and largely recover the original image content.
 
  \begin{table}
  \caption{Performance of our method with feedback loops under adversarial attacks on CIFAR-10 dataset.}
  \label{feedback_loop}
  \centering
  \begin{tabular}{c|ccccc}
    \toprule
     Attack Method & Gen1 & Gen2 & Gen3 &Gen4  \\
    \midrule
    FGS & 57.64\% & 78.04\% & 78.15\% & 78.31\% \\
    PGD & 78.46\%&85.36\%&86.25\%&86.55\%   \\
    BPDA & 19.40\%&79.12\%&79.28\%&79.79\%  \\
    
    \bottomrule
  \end{tabular}
\end{table}

 \begin{table}
  \caption{Performance analysis of algorithm components.}
  \label{deadzone-eval}
  \centering
  \begin{tabular}{c|ccc}
    \toprule
     Defense Methods & FGS & PGD & BPDA  \\
    \midrule
    Our Method (Full Alg.) & 88.51\% & 88.04\% & 85.77\%  \\
    \quad - Without Transform & 79.32\%  & 79.35\%  & 79.62\%   \\
   \quad - Without Feedback & 77.12\%  & 78.46\%  & 19.37\%   \\
    \bottomrule
  \end{tabular}
\end{table}

\textbf{(3) In-depth analysis of major algorithm components.}
In the following ablation studies, we perform in-depth analysis of major  components of our EGC-FL algorithm, which includes the transform, deadzone, and the EGC network with feedback loops. 
In Table \ref{deadzone-eval}, the first row shows the classification accuracy of images after defense with our proposed EGC-FL method (full algorithm) on the CIFAR-10 dataset with FGS, PGD, and BPDA attacks. The second row shows results without the transform. We can see that the accuracy drops about 7-9\%. The transform module is important because it can help protecting the original content from being damaged by the deadzone activation function $\delta(x)$ by aggregating the energy of the original image into a small number of large transform coefficients. 
The third row shows the results without the feedback loop. We can see  that it drops the accuracy by 10-11\% under the FGS and PGD attacks. For the powerful BPDA attack, the drop is very dramatic, about 66\%. With multiple feedback loops for progressive attack noise removal and original image reconstruction, it can significantly improve the defense performance, especially under powerful BPDA attacks. 

\begin{figure}
\begin{center}
\includegraphics[width=\linewidth]{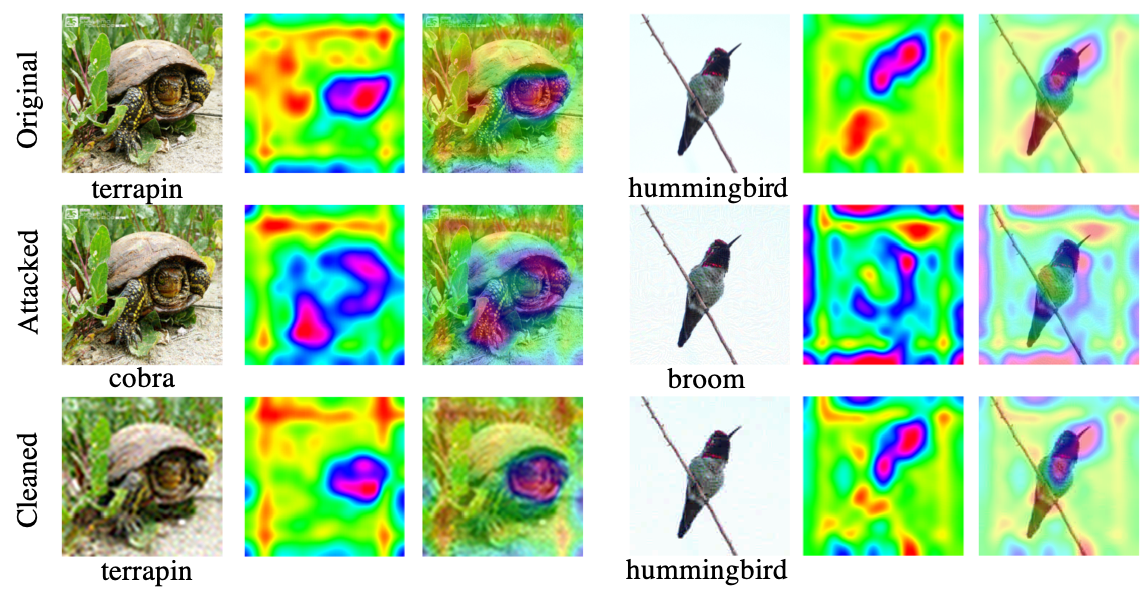}
\end{center}
\caption{Each pair of examples are feature maps corresponding to clean images (top), to their adversarial perturbed images (middle) and to their reconstructed images (bottom).}
\label{fig:feature_map}
\end{figure}

\textbf{(4) Visualizing the defense process.}
Network defense is essentially a denosing process of the feature maps. 
To further understand how the the proposed EGC-FL method works, we visualize the feature maps of original, attacked, and EGC-FL cleaned images. We use the feature map from the activation layer, the third from the last layer in the network.  
Figure \ref{fig:feature_map} shows two examples. In the first example, the first row is the original image (classified into terrapin), its gradient-weighted class activation heatmap, and the heatmap overlaid on the original image. 
The heatmap shows which parts of the original image the classification network is paying attention to. 
The second row shows the attacked image (being classified into cobra), heatmap, and the heatmap overlaid on the attacked image. 
We can see that the feature map is very noisy and the heatmap is distorted. 
The third row shows the EGC-cleaned images. We can see that both the feature map and heatmaps have been largely restored.

\section{Conclusion}
We have developed a new method for defending deep neural networks against adversarial attacks based on the EGC-FL network. 
This network is able to recover the original image while cleaning up the residual attack noise. 
We introduced a transformed deadzone layer into the defense network, which consists  of  an  orthonormal  transform  and  a deadzone-based activation function, to destroy the sophisticated  noise  pattern  of  adversarial  attacks. 
By constructing a generative cleaning network with a feedback loop, we are able to generate an ensemble of  diverse estimations of the original clean image.  We then learned a network to fuse this set of diverse estimation images together  to  restore  the  original image.
Our extensive experimental results demonstrated that our approach outperforms the state-of-art methods by large margins in both white-box and black-box attacks. Our ablation studies have demonstrated that the major components of our method, the transformed deadzone layer and the ensemble generative cleaning network with feedback loops, are both critical, contributing significantly to the overall performance.

\section*{Acknowledgement}
This work was supported in part by National Science Foundation under grants 1647213 and 1646065.
Any opinions, findings, and conclusions or recommendations expressed in this material are those of the authors and do not necessarily reflect the views of the National Science Foundation.

{\small
\bibliographystyle{ieee_fullname}
\bibliography{egbib}
}

\end{document}